\documentclass[11pt]{article}

\usepackage{acl}
\usepackage{times}
\usepackage{latexsym}
\usepackage[T1]{fontenc}
\usepackage[utf8]{inputenc}
\usepackage{microtype}
\usepackage{inconsolata}
\usepackage{graphicx}
\usepackage{booktabs}
\usepackage{amsmath}
\usepackage{amssymb}
\usepackage{array}
\usepackage{enumitem}
\usepackage{multirow}
\usepackage{tabularx}
\usepackage{xspace}

\newcommand{\method}{\textsc{CounterRefine}\xspace}
\newcommand{\baseline}{\textsc{Baseline-RAG}\xspace}
\newcommand{\KEEP}{\texttt{KEEP}\xspace}
\newcommand{\REVISE}{\texttt{REVISE}\xspace}
\newcolumntype{L}[1]{>{\raggedright\arraybackslash}p{#1}}
\newcolumntype{Y}{>{\raggedright\arraybackslash}X}

\title{CounterRefine: Answer-Conditioned Counterevidence Retrieval for Inference-Time Knowledge Repair in Factual Question Answering}

\author{
Tianyi Huang\thanks{Corresponding author.} \\
Ryquo \\
\texttt{tianyi@ryquo.com}
\And
Ying Kai Deng \\
App-In Club \\
\texttt{kai.deng@appinclub.org}
}

\begin{document}
\maketitle

\begin{abstract}
In factual question answering, many errors are not failures of access but failures of commitment: the system retrieves relevant evidence, yet still settles on the wrong answer. We present \method, a lightweight repair layer for short-form RAG that treats the first answer as a hypothesis to test. Given a draft, \method issues answer-conditioned expansion queries to retrieve candidate-specific evidence, then applies a constrained \KEEP or \REVISE refinement step whose proposed revisions are accepted only after deterministic validation. The design is intentionally narrow: it adds one evidence-gathering pass and one guarded refinement call rather than replacing the retriever or building a broad agentic system. On the full SimpleQA benchmark, \method improves a matched one-pass RAG baseline by up to 5.8 correct-rate points; in the full Claude trace, it changes only 5.6\% of outputs, with 180 beneficial outcome changes and 8 harmful ones. These findings suggest a simple but important direction for knowledgeable foundation models: beyond accessing evidence, they should also be able to use that evidence to reconsider and, when necessary, repair their own answers.
\end{abstract}

\section{Introduction}

Large language models are increasingly used as knowledge interfaces, but a central difficulty remains unresolved: a model can retrieve relevant material and still commit to the wrong fact. In short-form factual QA, these errors are unforgiving. A wrong year, nearby entity, or almost-right title is still fully incorrect. This makes factual QA a useful setting for studying a broader question that matters for knowledgeable foundation models: when an answer is wrong, can the system repair it at inference time rather than only during pre-training or parameter editing?

Retrieval-augmented generation (RAG) improves factuality by grounding generation in external evidence rather than relying only on parametric memory \citep{lewis2020rag,karpukhin2020dpr}. Yet retrieval alone does not eliminate candidate-selection errors. A first-pass retriever is optimized for \emph{topic relevance}, not necessarily for \emph{candidate discrimination}. Once a draft answer exists, however, the retrieval problem changes. A useful next query may include the candidate answer itself: if the draft year is wrong, adding that year to the query can surface evidence that falsifies it; if the draft entity is ambiguous, answer-conditioned re-querying can reveal a more discriminative title, list entry, or sentence.

We build on this observation with \method, a retrieval-based short-answer system that first drafts an answer from web snippets and then performs one answer-conditioned evidence-gathering pass. This second stage is best viewed as a focused form of query expansion, not as a new general-purpose retrieval algorithm. Its role is to expose evidence that bears directly on a concrete hypothesis. A constrained refinement model then decides whether to \KEEP the draft or \REVISE it, and a deterministic validator blocks unsupported or ill-formed rewrites. The result is a simple repair layer that can sit on top of an existing retrieval pipeline.

Our empirical evidence focuses on matched evaluation. On the full 4,326-question SimpleQA benchmark with the official grader \citep{wei2024simpleqa,openai2025simpleevals}, \method improves correct rate by +4.0 points with Claude Sonnet 4.6 and +5.8 points with GPT-5 over the corresponding one-pass retrieval baselines. We also performed preliminary transfer checks on 100 and 300 sample slices of HotpotQA distractor development \citep{yang2018hotpotqa}, where the exact match was improved for both backbones. These results support a bounded claim: answer-conditioned evidence gathering can be useful when it is paired with a conservative gate, but it is not a replacement for broader verification systems or stronger retrieval.

Our contributions are: (1) we instantiate answer-conditioned query expansion as a candidate-testing step inside a guarded short-answer repair layer; (2) we show that the complete layer improves over a matched retrieval baseline under the official SimpleQA evaluation pipeline; and (3) we provide an intervention profile and representative examples that clarify when the method helps and when it can still fail.

\section{Related Work}

\paragraph{Retrieval-grounded factual QA.}
Retrieval-augmented generation improves knowledge-intensive NLP by grounding generation in external documents \citep{lewis2020rag}. Dense Passage Retrieval further improved open-domain QA with learned neural retrieval \citep{karpukhin2020dpr}. More recent work has explored adaptive retrieval and critique, for example by teaching models when to retrieve and how to reflect on retrieved evidence \citep{asai2024selfrag}. Other work has targeted robustness to noisy retrieval \citep{yu2024chainnote} or refreshed model knowledge through search-engine augmentation \citep{vu2024freshllms}. Our method does not redesign the retriever or the full RAG stack; it adds a small second-pass evidence-gathering step that is conditioned on the first answer.

\paragraph{Query expansion and answer-conditioned retrieval.}
The retrieval step in \method is closely related to query expansion. Classical query expansion and pseudo-relevance-feedback methods have long expanded queries using terms inferred from local document evidence, global corpus statistics, or estimated relevance models \citep{xu1996localglobal_sigir,lavrenko2001rlm_sigir,carpineto2012aqe_survey}. Recent LLM-based variants generate auxiliary text for retrieval, including HyDE, which retrieves through hypothetical document embeddings \citep{hyde2023acl_query}, and Query2Doc, which expands queries with LLM-generated pseudo-documents \citep{query2doc2023emnlp_query}. \method shares the broad idea of using auxiliary generated text to improve retrieval. Its focus is narrower: the auxiliary text is the system's own draft answer, and the expanded retrieval pass is embedded in a conservative \KEEP/\REVISE repair layer for short factual answers.

\paragraph{Verification and self-correction.}
A broad line of work improves factuality by checking or revising an initial answer. Chain-of-Verification generates verification questions before producing a final answer \citep{dhuliawala2024chain}. CRITIC uses tool-interactive critique to validate and amend outputs \citep{gou2024critic}. RARR retrieves evidence and revises unsupported claims while trying to preserve the original text \citep{gao2023rarr}. SelfCheckGPT detects likely hallucinations through disagreement across samples \citep{manakul2023selfcheckgpt}. \method is closest in spirit to this family. We do not claim that it dominates these broader systems; rather, we study a deliberately constrained setting: one additional evidence-gathering phase, one refinement call, and a short-answer decision with deterministic validation.

\paragraph{Knowledge repair and model editing.}
Wrong knowledge can also be addressed by editing model parameters directly. ROME and MEMIT show that factual associations in language models can be modified through targeted weight updates \citep{meng2022rome,meng2023memit}. Those methods operate on the model's stored knowledge itself. \method is complementary: it performs answer-level repair at inference time instead of changing model parameters, using external evidence to challenge and revise a candidate answer.

\paragraph{Benchmarks and conflicting evidence.}
Factuality has been studied through claim verification datasets such as FEVER \citep{thorne2018fever}, factual QA benchmarks such as TruthfulQA \citep{lin2022truthfulqa}, long-form factuality metrics such as FACTSCORE \citep{min2023factscore}, and broader surveys of LLM factuality \citep{wang2024factuality}. SimpleQA is particularly relevant here because it isolates short-form factual precision in a benchmark of 4,326 questions with a single indisputable answer \citep{wei2024simpleqa}. Recent work has also emphasized that retrieval errors are not limited to irrelevant documents; systems often face ambiguous, noisy, or conflicting evidence \citep{wang2025conflicting}. Our setting is simpler than that literature's multi-document conflict benchmarks, but the motivating issue is related: even in short-answer QA, retrieved snippets can weakly support multiple nearby candidates. Answer-conditioned re-querying is one lightweight way to make those conflicts visible before finalizing an answer.

\begin{figure*}[t]
\centering
\includegraphics[width=\textwidth]{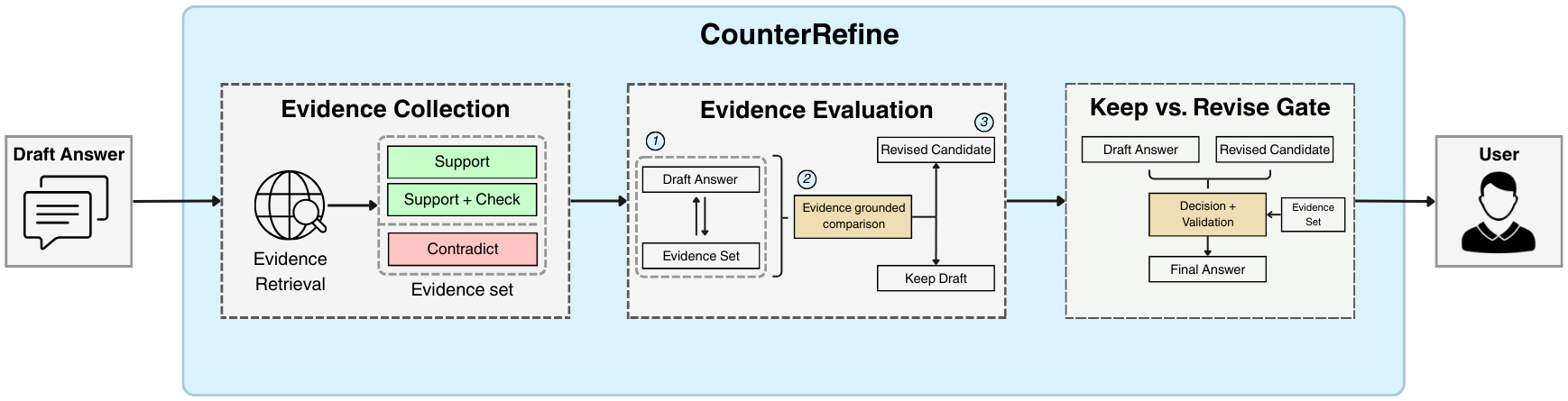}
\caption{Schematic overview of \method. Starting from a draft answer produced by \baseline, \method collects additional evidence intended to support, check, or contradict that candidate, evaluates the draft answer against the resulting evidence set, and then passes the result to a constrained keep-vs.-revise gate. The final answer is either the original draft or a revised candidate accepted under evidence-based validation using the same evidence set. For readability, the figure abstracts away some implementation details from the text, including question-type-conditioned query selection and per-query snippet aggregation.}
\label{fig:method}
\end{figure*}

\section{Method}

\subsection{Overview}

We refer to the first-pass retrieval system as \baseline. Given a question $q$, \method first runs \baseline to produce an initial answer $a_0$, then runs one answer-conditioned refinement stage:
\begin{align}
R_0 &= \mathrm{Retrieve}(q, k_b), \\
a_0 &= f_{\text{draft}}(q, R_0), \\
R_1 &= \mathrm{RetrieveRefine}(q, a_0, R_0), \\
a^\star &= f_{\text{refine}}(q, a_0, R_1).
\end{align}
The pipeline is designed as a focused reliability layer on top of an existing retrieval system.

\subsection{Stage 1: retrieval-based drafting}

The baseline stage retrieves up to $k_b$ evidence snippets for the original question:
\begin{equation}
R_0 = \mathrm{Retrieve}(q, k_b).
\end{equation}
For the SimpleQA web-retrieval experiments, $k_b = 5$. For the HotpotQA component ablation, the local distractor-context runner keeps up to 10 first-pass paragraphs. The retriever returns tuples of title, URL, and evidence text when URLs are available, and duplicates are removed before prompt construction.

The drafting model answers using only the retrieved evidence:
\begin{equation}
a_0 = f_{\text{draft}}(q, R_0).
\end{equation}
The draft prompt enforces a short answer and disallows hedging or explanation. If retrieval fails to yield a valid answer, the code falls back to a closed-book short-answer prompt; if that also fails, a minimal type-dependent default is emitted.

\subsection{Stage 2: answer-conditioned query expansion}

The refinement queries depend on the draft answer. Let $t(q)$ denote a coarse question type inferred from the question string. The refinement query set is
\begin{equation}
Q(q,a_0)=\{q,\; q \Vert a_0\} \cup
\begin{cases}
\{a_0\}, & t(q)\in\mathcal{T},\\
\varnothing, & \text{otherwise}
\end{cases}
\end{equation}
where $\Vert$ denotes concatenation of the question and the surface form of the draft answer with a separator, and
\[
\mathcal{T}=\{\texttt{who},\texttt{where},\texttt{when},\texttt{year},\texttt{number}\}.
\]
Thus, every example receives an answer-conditioned query containing both the original question and the draft answer, and slot-like questions also receive a bare-answer query. This is intended to retrieve evidence that directly bears on the provisional answer, including evidence that may support, sharpen, or contradict it.

The question-type heuristic is intentionally simple. It labels yes/no questions when the question begins with an auxiliary verb; year questions when the question contains cues such as \emph{what year} or \emph{in what year}; temporal questions when it begins with \emph{when} or asks for a month, date, or timeframe; numeric questions when it contains cues such as \emph{how many}, \emph{how much}, \emph{population}, or \emph{number}; location questions when it begins with \emph{where} or asks for a city, county, town, village, municipality, or neighborhood; and person questions when it begins with \emph{who}, \emph{whom}, or \emph{whose}. All other questions are assigned to a residual \texttt{other} type.

For each $q' \in Q(q,a_0)$, the system retrieves up to $k_r$ evidence snippets, with $k_r = 5$ in the current implementation, and merges them with the baseline support set:
\begin{equation}
\begin{aligned}
R_1
&= \mathrm{Dedupe}\Bigl(
R_0 \cup \bigcup_{q' \in Q(q,a_0)}
\mathrm{Retrieve}(q', k_r)
\Bigr).
\end{aligned}
\end{equation}

\begin{table}[t]
\centering
\small
\setlength{\tabcolsep}{4pt}
\renewcommand{\arraystretch}{1.12}
\begin{tabularx}{\columnwidth}{L{1.7cm}Y}
\toprule
Step & Example trace \\
\midrule
Question & Who was awarded the Oceanography Society's Jerlov Award in 2018? \\
Draft & Collin Roesler \\
Second-pass queries & Original question; original question + ``Collin Roesler''; ``Collin Roesler'' \\
Retrieved signal & Award-list evidence names Annick Bricaud as the 2018 recipient. \\
Decision & \REVISE to Annick Bricaud \\
\bottomrule
\end{tabularx}
\caption{Illustrative Stage-2/Stage-3 trace. The second pass does not solve the question from scratch; it tests a concrete draft answer against newly surfaced evidence.}
\label{tab:example}
\end{table}

Table~\ref{tab:example} gives a compact example. The first-pass answer is plausible but wrong. Conditioning retrieval on the draft answer surfaces a more discriminative award-list signal, after which the refinement gate can revise the answer.

\subsection{Stage 3: constrained refinement}

The refiner receives the question, the baseline answer, and the merged evidence set, and must output exactly three fields:
\begin{quote}
\small
\texttt{DECISION: KEEP or REVISE}\\
\texttt{ANSWER: <short answer only>}\\
\texttt{EVIDENCE: <one short evidence snippet or NONE>}
\end{quote}
The prompt instructs the model to \KEEP when the baseline is already the best supported answer and to \REVISE only when the evidence strongly supports a different or more exact short answer. The prompt does not ask for a chain of reasoning; the evidence field is used only as a short grounding signal for validation.

Let $(d, \hat{a}, e)$ denote the parsed refinement output. If $d = \texttt{KEEP}$, the system returns $a_0$. If $d = \texttt{REVISE}$, the proposed answer must pass deterministic validation before it is accepted. Malformed outputs, missing required fields, empty answers, and non-answer strings are treated conservatively: the system falls back to the draft rather than emitting an unvalidated rewrite.

\subsection{Deterministic validation and canonicalization}

The validator blocks revisions that are unsupported, ill-typed, or stylistically incompatible with short-answer evaluation. A proposed rewrite is rejected if any of the following hold:
\begin{enumerate}[leftmargin=1.2em]
    \item it is empty, non-responsive, or identical to the draft after normalization;
    \item for yes/no questions, it is not exactly \texttt{yes} or \texttt{no};
    \item for entity-style questions, it is clause-like, overly long, or begins with a descriptor phrase;
    \item for temporal or numeric questions, it lacks an explicit temporal or numeric marker;
    \item the refiner provides no evidence snippet;
    \item lexical grounding between the revised answer and the cited evidence is too weak.
\end{enumerate}
Rule 6 is implemented as a necessary grounding check after lowercasing and punctuation removal. For temporal and numeric questions, at least one year, digit-bearing token, or month token from the proposed answer must appear in the cited evidence. For entity-style answers, at least one normalized non-stopword answer token must appear in the cited evidence; for multi-token answers, the check favors contiguous or near-contiguous overlap when available. This rule is intentionally conservative: it cannot prove that a revision is correct, but it blocks revisions that are not even lexically anchored in the evidence supplied by the refiner.

Accepted revisions are then canonicalized with question-type-specific rules, such as extracting a 4-digit year, compacting number spans, stripping leading prepositions for locations, or removing parenthetical descriptors from names. Only proposed revisions are validated; \KEEP decisions preserve the baseline answer unchanged. The final output is
\begin{equation}
a^\star =
\begin{cases}
\mathrm{Canon}(q,\hat{a}), & \text{if the revision is accepted,}\\
a_0, & \text{otherwise.}
\end{cases}
\label{eq:final}
\end{equation}

\section{Experimental Setup}

\subsection{Benchmarks and metrics}

We evaluate on two benchmarks. Our primary benchmark is SimpleQA, which contains 4,326 short, fact-seeking questions with a single indisputable answer \citep{wei2024simpleqa}. We use the official SimpleQA grader from the public \texttt{simple-evals} implementation \citep{openai2025simpleevals}. Following the official setup, we report the correct rate and F1 score.

As a preliminary transfer setting, we evaluate on 100-example and 300-example slices from the HotpotQA distractor development set \citep{yang2018hotpotqa}. HotpotQA is useful here because it is more multi-hop and span-oriented than SimpleQA. We report the standard exact match (EM) and token-overlap F1 metrics. We also run a 300-example HotpotQA component ablation with the same backbone within each run. This ablation is intended for within-configuration attribution rather than as a new benchmark-level HotpotQA claim.

\subsection{Systems and decoding}

Our primary full-benchmark run uses Claude Sonnet 4.6 for both drafting and refinement \citep{anthropic}. Retrieval is implemented through the OpenAI web search API in a simple RAG pipeline that returns web evidence for subsequent answer generation \citep{websearch, lewis2020rag}. The baseline system, \baseline, uses the same retrieval component and the same drafting backbone but does not run the answer-conditioned refinement stage. This keeps the comparison focused on the contribution of \method rather than on backbone or first-pass retrieval differences.

We also ran matched experiments with GPT-5 on SimpleQA and HotpotQA \citep{openai}. All reported numbers are single-run benchmark results under the same retrieval and prompting configuration within each backbone. Where the provider API exposes a temperature or comparable stochasticity control, we use a low-randomness setting. We do not report multi-seed variance; this is one reason we emphasize the full SimpleQA matched-baseline comparison over small-slice differences.

\subsection{Computational profile}

Table~\ref{tab:cost} reports per-question operation counts for the SimpleQA web-retrieval setting before deduplication. We report counts rather than dollar cost because provider pricing, caching, and latency vary by model, date, and deployment environment. The main overhead is bounded and explicit: \method adds one refinement model call and two or three second-pass retrieval queries, depending on question type.

\begin{table}[t]
\centering
\small
\setlength{\tabcolsep}{2.5pt}
\renewcommand{\arraystretch}{1.15}

\begingroup
\renewcommand{\tabularxcolumn}[1]{m{#1}}
\begin{tabularx}{\columnwidth}{@{}
  >{\raggedright\arraybackslash}m{0.34\columnwidth}
  *{3}{>{\centering\arraybackslash}X}
@{}}
\toprule
System & \shortstack{Retrieval\\queries} & \shortstack{Raw\\snippets} & \shortstack{Model\\calls} \\
\midrule
\baseline & 1 & $\leq 5$ & 1 draft \\
\specialrule{0.2pt}{4pt}{5pt}
\method & 3--4 & $\leq 15$--$20$ & \shortstack{1 draft +\\1 refine} \\
\bottomrule
\end{tabularx}
\endgroup

\caption{Per-question computational profile for the SimpleQA web-retrieval setting before deduplication. The 3--4 retrieval-query range for \method includes the shared first-pass query and depends on whether the question type triggers the bare-answer query.}
\label{tab:cost}
\end{table}

\section{Results}

\subsection{Performance across benchmarks and backbones}

\begin{table*}[t]
\centering
\small
\setlength{\tabcolsep}{4.2pt}
\renewcommand{\arraystretch}{1.12}
\begin{tabular}{llcccccc}
\toprule
\multirow{2}{*}{Benchmark} & \multirow{2}{*}{Metric} & \multicolumn{3}{c}{Claude 4.6} & \multicolumn{3}{c}{GPT-5} \\
\cmidrule(lr){3-5}\cmidrule(lr){6-8}
& & Base-RAG & \method & $\Delta$ & Base-RAG & \method & $\Delta$ \\
\midrule
\multirow{2}{*}{SimpleQA (100)}
  & Correct$\uparrow$ & 65.0 & \textbf{75.0} & +10.0 & 57.0 & \textbf{71.0} & +14.0 \\
  & F1$\uparrow$      & 65.0 & \textbf{75.0} & +10.0 & 59.7 & \textbf{73.2} & +13.5 \\
\midrule
\multirow{2}{*}{SimpleQA (4,326)}
  & Correct$\uparrow$ & 63.7 & \textbf{67.7} & +4.0 & 67.3 & \textbf{73.1} & +5.8 \\
  & F1$\uparrow$      & 64.1 & \textbf{68.1} & +4.0 & 58.6 & \textbf{72.1} & +13.5 \\
\midrule
\multirow{2}{*}{HotpotQA (100)}
  & EM$\uparrow$ & 74.0 & \textbf{79.0} & +5.0 & 71.0 & \textbf{76.0} & +5.0 \\
  & F1$\uparrow$ & 82.2 & \textbf{84.9} & +2.7 & 83.7 & \textbf{84.5} & +0.8 \\
\midrule
\multirow{2}{*}{HotpotQA (300)}
  & EM$\uparrow$ & 70.0 & \textbf{76.0} & +6.0 & 68.0 & \textbf{71.7} & +3.7 \\
  & F1$\uparrow$ & 83.9 & \textbf{86.1} & +2.2 & 83.6 & \textbf{84.9} & +1.3 \\
\bottomrule
\end{tabular}
\caption{Main results across backbones and benchmarks. \baseline is the matched one-pass retrieval baseline using the same retriever and backbone as \method. SimpleQA uses the official grader. HotpotQA uses the distractor development setting on 100-example and 300-example slices. The HotpotQA (300) \method values are the full-system row of Table~\ref{tab:hotpot_ablation}. All values are percentages.}
\label{tab:main}
\end{table*}

Table~\ref{tab:main} reports the main quantitative results. Adding \method on top of \baseline improves the primary exactness-oriented metric in each reported setting. The strongest evidence is the full SimpleQA evaluation, where both systems use the same retriever and backbone. The smaller SimpleQA and HotpotQA slices are useful checks, but they should not be read as establishing broad benchmark-level superiority.

\begin{figure}[t]
    \centering
    \includegraphics[width=\columnwidth]{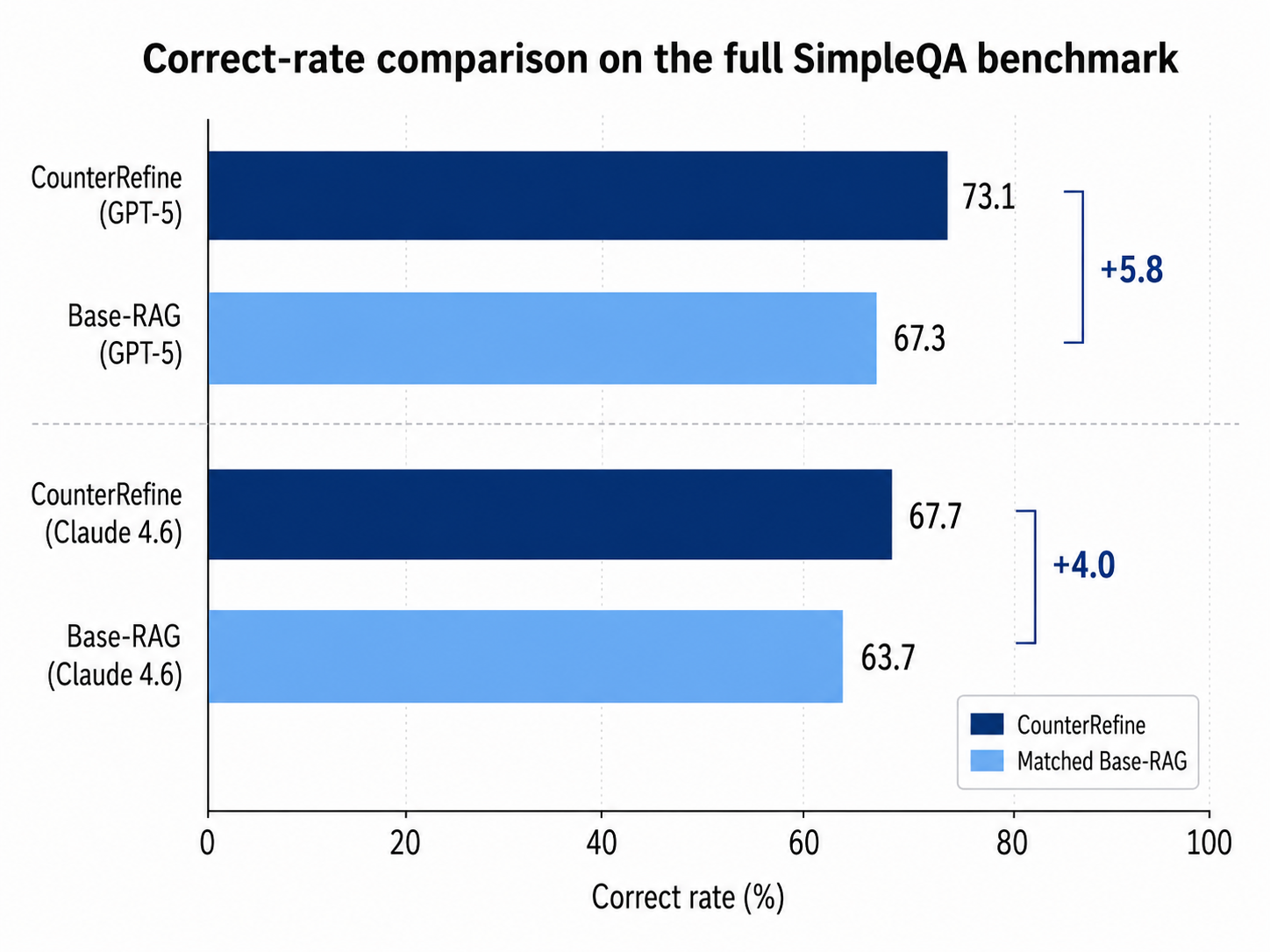}
    \caption{Correct-rate comparison on the full SimpleQA benchmark. \method is compared only with the matched one-pass retrieval baseline using the same backbone and retrieval setup.}
    \label{fig:main_result}
\end{figure}

Two aspects of the result are worth noting. First, the improvement appears on top of an already retrieval-grounded baseline rather than a no-retrieval model. The comparison is therefore between a matched one-pass retrieval pipeline and the same pipeline augmented with answer-conditioned refinement. Second, the gains are strongest on metrics that reward final answer exactness. This is most visible in the SimpleQA correct rate and HotpotQA EM, which matches the intended role of \method as a repair layer for final factual precision rather than a general-purpose reasoning scaffold.

For GPT-5 on full SimpleQA, the F1 gain is larger than the correct-rate gain. We interpret this cautiously. F1 is sensitive to answer surface form and tokenization, while \method canonicalizes accepted revisions into shorter slot-like answers. We therefore treat official SimpleQA correct rate as the primary metric and F1 as a secondary surface-overlap diagnostic.

The cross-benchmark pattern is also informative but preliminary. On SimpleQA, where a near-miss entity, date, or number is scored as fully wrong, answer-conditioned refinement yields clear gains under the official evaluation pipeline. On HotpotQA, the same mechanism improves exact match on the reported slices, including the 300-example runs aligned with the component ablation. Because these HotpotQA runs cover slices rather than the full development set, we use them as evidence of transfer potential rather than as a complete characterization of multi-hop performance.

\subsection{Component ablations on HotpotQA}

Table~\ref{tab:hotpot_ablation} reports a 300-example HotpotQA component ablation. The rows compare the full system with four nearby variants: a second pass that reuses only the original question, answer-conditioned refinement without deterministic validation, answer-conditioned refinement without canonicalization, and a simple reconsideration baseline with the same additional model-call budget. The Full \method row is the same HotpotQA-300 run reported in Table~\ref{tab:main}, using the same matched baseline. These rows are meant to compare variants within a fixed local distractor-context setup; they are not intended to replace the full SimpleQA evaluation as the main evidence for the method.

\begin{table*}[t]
\centering
\small
\setlength{\tabcolsep}{4.5pt}
\renewcommand{\arraystretch}{1.12}
\begin{tabular}{lcccc}
\toprule
\multirow{2}{*}{Variant} & \multicolumn{2}{c}{Claude 4.6} & \multicolumn{2}{c}{GPT-5} \\
\cmidrule(lr){2-3}\cmidrule(lr){4-5}
& EM$\uparrow$ & F1$\uparrow$ & EM$\uparrow$ & F1$\uparrow$ \\
\midrule
Full \method & 76.0 & 86.1 & 71.7 & 84.9 \\
Second pass with original question only & 76.0 & 86.1 & 71.7 & 84.9 \\
Answer-conditioned, no validator & 74.7 & 86.0 & 72.0 & 85.0 \\
Answer-conditioned, no canonicalization & 75.0 & 86.5 & 71.7 & 84.9 \\
Simple reconsideration & 74.7 & 86.0 & 72.0 & 85.0 \\
\bottomrule
\end{tabular}
\caption{HotpotQA-300 component ablation. The table compares the full \method pipeline with variants that use only the original question in the second pass, remove deterministic validation, remove answer canonicalization, or replace the guarded refinement step with simple reconsideration. Values are percentages; the Full \method row corresponds to the HotpotQA-300 system row in Table~\ref{tab:main}, and EM is rounded from exact-match counts over 300 examples.}
\label{tab:hotpot_ablation}
\end{table*}

The ablation results are useful but limited. Relative to the matched HotpotQA-300 baseline in Table~\ref{tab:main}, the full system improves Claude 4.6 by +6.0 EM and +2.2 F1, and GPT-5 by +3.7 EM and +1.3 F1. The nearby variants fall in a narrow range, so this slice does not isolate a large independent contribution from any single component. We use the ablation as a within-configuration check on the second-pass repair family; the strongest evidence for the guarded design remains the full SimpleQA matched-baseline gain and selective-intervention profile.

\subsection{Intervention profile}

The full Claude SimpleQA trace helps explain the matched-baseline gain. \method changes only a small fraction of predictions, and those changes are mostly beneficial: the system revises 5.6\% of examples, helps 180, hurts 8, and produces zero refiner-format failures. Beneficial outcome-changing revisions therefore outnumber harmful ones by 22.5 to 1. Here, a refiner-format failure means an API/runtime failure or malformed output that prevents parsing the required fields. Validator-rejected \REVISE proposals are handled conservatively by returning the draft and are separate from this failure count. The not-attempted rate remains fixed at 1.2\%, so the improvement is not driven by abstaining more often.

\begin{figure}[h]
    \centering
    \includegraphics[width=\columnwidth]{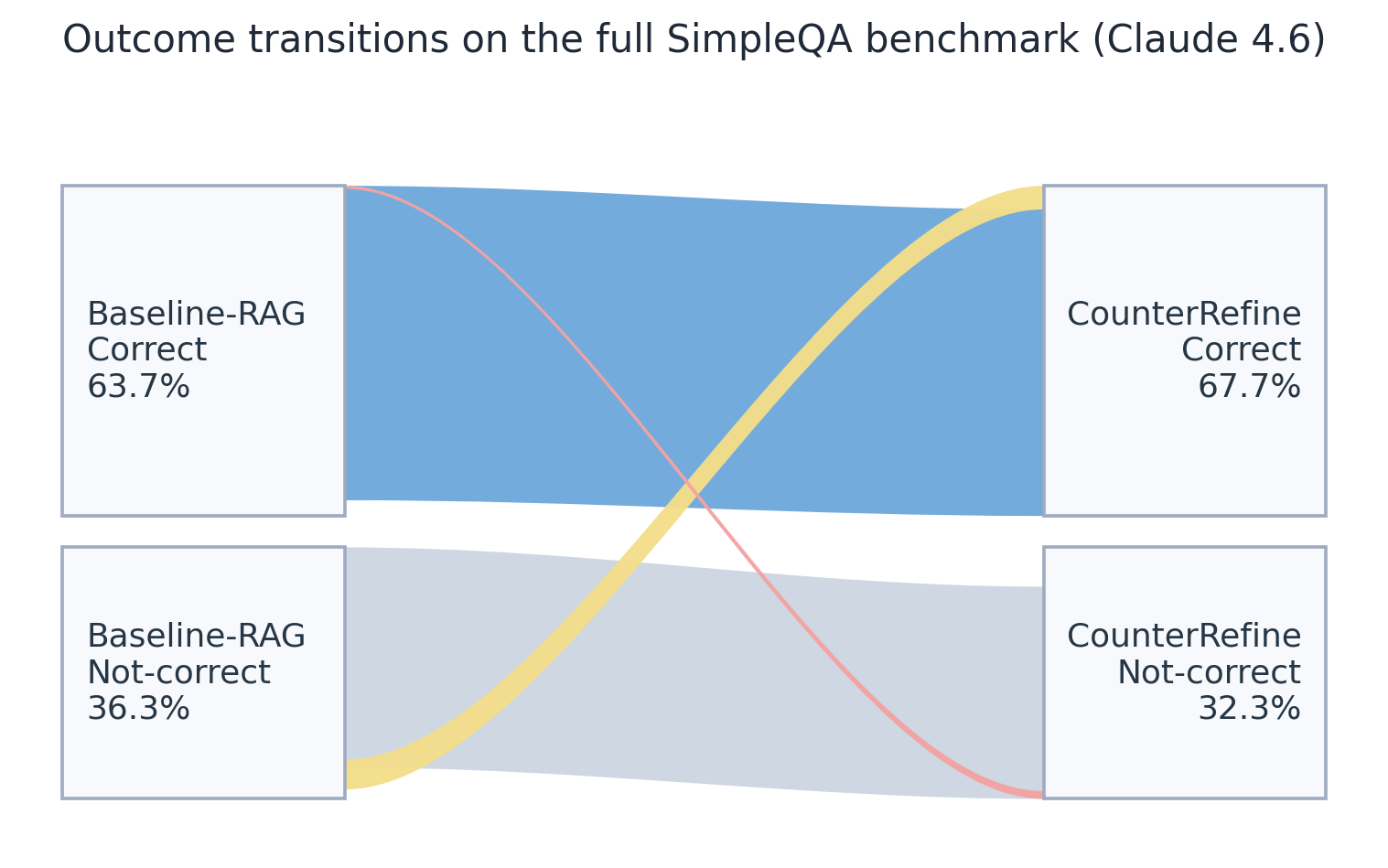}
    \caption{Outcome transitions on the full SimpleQA benchmark for Claude 4.6. Of all examples, 63.5\% stay correct, 32.1\% stay not-correct, 4.2\% are corrected (incorrect $\rightarrow$ correct), and 0.2\% are harmed (correct $\rightarrow$ incorrect).}
    \label{fig:sankey}
\end{figure}

Figure~\ref{fig:sankey} makes the structure of the gain explicit. Most examples are unchanged, but among changed outcomes, incorrect $\rightarrow$ correct transitions dominate correct $\rightarrow$ incorrect ones by a wide margin. This is the intended behavior for a post-retrieval repair layer: it should intervene selectively, and when it does, it should usually move the answer in the right direction.

\subsection{Representative failures and successes}

\begin{table*}[t]
\centering
\small
\setlength{\tabcolsep}{4.5pt}
\renewcommand{\arraystretch}{1.18}
\begin{tabularx}{\textwidth}{L{5.0cm}L{1.9cm}L{1.9cm}L{1.9cm}Y}
\toprule
Question (abridged) & \baseline & Final & Gold & Diagnosis \\
\midrule
\multicolumn{5}{l}{\textbf{Representative hurt cases}} \\
Who appointed the Chief Justice of India, Mirza Hameedullah Beg, in 1977? & Fakhruddin Ali Ahmed & Indira Gandhi & Fakhruddin Ali Ahmed & Nearby evidence named the government behind the elevation rather than the appointing president requested by the question. \\
\addlinespace[0.35em]
What day, month, and year was the municipality of Arboletes, Antioquia, Colombia, founded? & 1920 & August 1958 & July 20, 1920 & The refinement evidence referred to municipal establishment rather than the original founding date. \\
\addlinespace[0.35em]
On what day, month, and year did Manny Pacquiao marry Jinkee Jamora? & May 10, 1999 & May 10, 2000 & May 10, 1999 & A noisy timeline-style snippet overrode an already correct baseline year. \\
\midrule
\multicolumn{5}{l}{\textbf{Representative helped cases}} \\
Who was awarded the Oceanography Society's Jerlov Award in 2018? & Collin Roesler & Annick Bricaud & Annick Bricaud & Answer-conditioned retrieval surfaced an award-list entry explicitly naming the 2018 winner. \\
\addlinespace[0.35em]
What were the day, month, and year of death of Mehr Chand Mahajan? & 5 December 1984 & 11 December 1967 & 11 December 1967 & The refinement stage found a biographical snippet containing the exact parenthetical death date. \\
\addlinespace[0.35em]
In which year did Fazal Ilahi Chaudhry join the Muslim League? & 1940s & 1942 & 1942 & Re-querying with the draft answer surfaced the exact year instead of a decade-level answer. \\
\bottomrule
\end{tabularx}
\caption{Representative failures and successes from the full Claude trace log.}
\label{tab:cases}
\end{table*}

Table~\ref{tab:cases} illustrates the main failure and success patterns. The hurt cases are not random; they are mostly cases of \emph{relation confusion} or \emph{event mismatch}. In these examples, the refinement stage retrieves evidence about a nearby but different relation, milestone, or date, and that adjacent fact is plausible enough to override an initially correct answer. This suggests that the main remaining weakness is not lack of evidence, but insufficient discrimination between closely related factual relations.

The helped cases show the complementary pattern. \baseline is often nearly correct but too coarse, under-specified, or anchored on the wrong nearby candidate. Once the draft answer is injected back into retrieval, the second-stage search can surface a decisive snippet that the first pass failed to isolate. The benefit is therefore not an unconstrained increase in reasoning effort; it is a change in what evidence becomes visible after a concrete hypothesis has been formed.

\section{Discussion}

The results support a bounded interpretation: one-pass retrieval sometimes fails not because relevant evidence is globally unavailable, but because the system commits to a plausible candidate before retrieving evidence that would better test that candidate. \method addresses this failure mode by changing the role of retrieval after drafting. The second retrieval pass is no longer only about finding topic-relevant material; it is also about checking a concrete factual hypothesis.

This framing also clarifies the scope of the claim. The answer-conditioned retrieval step is a form of query expansion, and Table~\ref{tab:main} evaluates the complete repair layer rather than query expansion alone. The HotpotQA ablations in Table~\ref{tab:hotpot_ablation} support a cautious reading: nearby second-pass variants perform similarly on that slice, so we do not claim that each component contributes a large independent gain in every setting. The main empirical claim is more modest: in short-form factual QA, a matched RAG system can benefit from a lightweight second-pass repair step when candidate-specific retrieval is paired with a conservative keep-or-revise gate.

\section{Conclusion}

We presented \method, a retrieval-based short-answer QA system that answers first and then retrieves evidence targeted at its own provisional answer. Across the reported evaluations, the method improves a matched one-pass RAG baseline, with the strongest evidence coming from full SimpleQA evaluation under the official grader and the HotpotQA results serving as preliminary transfer checks. The results suggest a simple repair pattern: treat the first answer as a hypothesis, retrieve evidence that tests that hypothesis, and accept revisions only when they pass conservative validation.

\section*{Limitations}

The empirical evidence is strongest for short-form factual settings, particularly SimpleQA, where exact entity, date, and value corrections are central to evaluation. The HotpotQA results are positive but cover limited slices rather than the full benchmark; they should be interpreted as preliminary evidence of transfer potential, not as a complete characterization of multi-hop performance.

Our primary comparison is a matched one-pass retrieval baseline. This isolates the value of adding \method to the same retriever and backbone, but it does not establish superiority over broader verification and editing systems such as RARR, CRITIC, Chain-of-Verification, or agentic tool-use pipelines. Our component ablations are limited to a 300-example HotpotQA slice and show small differences among nearby variants. They are useful as a diagnostic check, but they do not fully attribute the SimpleQA gains among answer-conditioned retrieval, deterministic validation, and canonicalization.

Finally, \method is designed to repair a draft answer after retrieval, not to replace retrieval itself or to solve long-form generation problems. Its effectiveness depends on whether retrieval surfaces evidence that is informative enough for the validator and refiner. We have not yet evaluated smaller refinement models or latency-optimized deployment variants, which are natural directions for reducing cost.

\section*{Broader Impact and Ethical Considerations}

\method points to a practical direction for improving retrieval-grounded QA systems: before an answer reaches a user, the system can use additional evidence to test whether a plausible draft should be kept or revised. Because the method is modular and compatible with standard retrieval pipelines, it may be useful as a reliability layer for search assistants, educational tools, and scientific interfaces where small factual errors can matter.

At the same time, inference-time repair can improve evidential support, but it does not guarantee truth. If retrieved evidence reflects social, historical, or institutional bias, a repair layer may reinforce those patterns rather than correct them. In high-stakes settings, repaired answers should therefore be treated as better-supported outputs rather than as definitive verification. Systems like \method should support, not replace, human judgment, especially in domains such as medicine, law, public policy, and education.

\bibliography{main}

@misc{lewis2020rag,
      title={Retrieval-Augmented Generation for Knowledge-Intensive NLP Tasks}, 
      author={Patrick Lewis and Ethan Perez and Aleksandra Piktus and Fabio Petroni and Vladimir Karpukhin and Naman Goyal and Heinrich Küttler and Mike Lewis and Wen-tau Yih and Tim Rocktäschel and Sebastian Riedel and Douwe Kiela},
      year={2021},
      eprint={2005.11401},
      archivePrefix={arXiv},
      primaryClass={cs.CL},
      url={https://arxiv.org/abs/2005.11401}, 
}

@misc{websearch,
  author       = {{OpenAI}},
  title        = {Web search},
  year         = {2026},
  howpublished = {\url{https://developers.openai.com/api/docs/guides/tools-web-search}},
  note         = {OpenAI API documentation}
}

@inproceedings{karpukhin2020dpr,
    title = "Dense Passage Retrieval for Open-Domain Question Answering",
    author = "Karpukhin, Vladimir  and
      Oguz, Barlas  and
      Min, Sewon  and
      Lewis, Patrick  and
      Wu, Ledell  and
      Edunov, Sergey  and
      Chen, Danqi  and
      Yih, Wen-tau",
    booktitle = "Proceedings of the 2020 Conference on Empirical Methods in Natural Language Processing (EMNLP)",
    month = nov,
    year = "2020",
    publisher = "Association for Computational Linguistics",
    url = "https://aclanthology.org/2020.emnlp-main.550/",
    doi = "10.18653/v1/2020.emnlp-main.550",
    pages = "6769--6781",
}

@inproceedings{asai2024selfrag,
title={Self-{RAG}: Learning to Retrieve, Generate, and Critique through Self-Reflection},
author={Akari Asai and Zeqiu Wu and Yizhong Wang and Avirup Sil and Hannaneh Hajishirzi},
booktitle={The Twelfth International Conference on Learning Representations},
year={2024},
url={https://openreview.net/forum?id=hSyW5go0v8}
}

@inproceedings{yu2024chainnote,
    title = "Chain-of-Note: Enhancing Robustness in Retrieval-Augmented Language Models",
    author = "Yu, Wenhao  and
      Zhang, Hongming  and
      Pan, Xiaoman  and
      Cao, Peixin  and
      Ma, Kaixin  and
      Li, Jian  and
      Wang, Hongwei  and
      Yu, Dong",
    booktitle = "Proceedings of the 2024 Conference on Empirical Methods in Natural Language Processing",
    month = nov,
    year = "2024",
    publisher = "Association for Computational Linguistics",
    url = "https://aclanthology.org/2024.emnlp-main.813/",
    doi = "10.18653/v1/2024.emnlp-main.813",
    pages = "14672--14685",
}

@inproceedings{vu2024freshllms,
    title = "{F}resh{LLM}s: Refreshing Large Language Models with Search Engine Augmentation",
    author = "Vu, Tu  and
      Iyyer, Mohit  and
      Wang, Xuezhi  and
      Constant, Noah  and
      Wei, Jerry  and
      Wei, Jason  and
      Tar, Chris  and
      Sung, Yun-Hsuan  and
      Zhou, Denny  and
      Le, Quoc  and
      Luong, Thang",
    booktitle = "Findings of the Association for Computational Linguistics: ACL 2024",
    month = aug,
    year = "2024",
    publisher = "Association for Computational Linguistics",
    url = "https://aclanthology.org/2024.findings-acl.813/",
    doi = "10.18653/v1/2024.findings-acl.813",
    pages = "13697--13720",
}

@inproceedings{dhuliawala2024chain,
    title = "Chain-of-Verification Reduces Hallucination in Large Language Models",
    author = "Dhuliawala, Shehzaad  and
      Komeili, Mojtaba  and
      Xu, Jing  and
      Raileanu, Roberta  and
      Li, Xian  and
      Celikyilmaz, Asli  and
      Weston, Jason",
    booktitle = "Findings of the Association for Computational Linguistics: ACL 2024",
    month = aug,
    year = "2024",
    publisher = "Association for Computational Linguistics",
    url = "https://aclanthology.org/2024.findings-acl.212/",
    doi = "10.18653/v1/2024.findings-acl.212",
    pages = "3563--3578",
}

@inproceedings{gao2023rarr,
    title = "{RARR}: Researching and Revising What Language Models Say, Using Language Models",
    author = "Gao, Luyu  and
      Dai, Zhuyun  and
      Pasupat, Panupong  and
      Chen, Anthony  and
      Chaganty, Arun Tejasvi  and
      Fan, Yicheng  and
      Zhao, Vincent  and
      Lao, Ni  and
      Lee, Hongrae  and
      Juan, Da-Cheng  and
      Guu, Kelvin",
    booktitle = "Proceedings of the 61st Annual Meeting of the Association for Computational Linguistics (Volume 1: Long Papers)",
    month = jul,
    year = "2023",
    publisher = "Association for Computational Linguistics",
    url = "https://aclanthology.org/2023.acl-long.910/",
    doi = "10.18653/v1/2023.acl-long.910",
    pages = "16477--16508",
}

@inproceedings{gou2024critic,
title={{CRITIC}: Large Language Models Can Self-Correct with Tool-Interactive Critiquing},
author={Zhibin Gou and Zhihong Shao and Yeyun Gong and yelong shen and Yujiu Yang and Nan Duan and Weizhu Chen},
booktitle={The Twelfth International Conference on Learning Representations},
year={2024},
url={https://openreview.net/forum?id=Sx038qxjek}
}

@inproceedings{manakul2023selfcheckgpt,
title={SelfCheck{GPT}: Zero-Resource Black-Box Hallucination Detection for Generative Large Language Models},
author={Potsawee Manakul and Adian Liusie and Mark Gales},
booktitle={The 2023 Conference on Empirical Methods in Natural Language Processing},
year={2023},
url={https://openreview.net/forum?id=RwzFNbJ3Ez}
}

@inproceedings{meng2022rome,
title={Locating and Editing Factual Associations in {GPT}},
author={Kevin Meng and David Bau and Alex J Andonian and Yonatan Belinkov},
booktitle={Advances in Neural Information Processing Systems},
editor={Alice H. Oh and Alekh Agarwal and Danielle Belgrave and Kyunghyun Cho},
year={2022},
url={https://openreview.net/forum?id=-h6WAS6eE4}
}

@inproceedings{meng2023memit,
title={Mass-Editing Memory in a Transformer},
author={Kevin Meng and Arnab Sen Sharma and Alex J Andonian and Yonatan Belinkov and David Bau},
booktitle={The Eleventh International Conference on Learning Representations },
year={2023},
url={https://openreview.net/forum?id=MkbcAHIYgyS}
}

@inproceedings{thorne2018fever,
    title = "{FEVER}: a Large-scale Dataset for Fact Extraction and {VER}ification",
    author = "Thorne, James  and
      Vlachos, Andreas  and
      Christodoulopoulos, Christos  and
      Mittal, Arpit",
    booktitle = "Proceedings of the 2018 Conference of the North {A}merican Chapter of the Association for Computational Linguistics: Human Language Technologies, Volume 1 (Long Papers)",
    month = jun,
    year = "2018",
    publisher = "Association for Computational Linguistics",
    url = "https://aclanthology.org/N18-1074/",
    doi = "10.18653/v1/N18-1074",
    pages = "809--819",
}

@misc{lin2022truthfulqa,
      title={TruthfulQA: Measuring How Models Mimic Human Falsehoods}, 
      author={Stephanie Lin and Jacob Hilton and Owain Evans},
      year={2022},
      eprint={2109.07958},
      archivePrefix={arXiv},
      primaryClass={cs.CL},
      url={https://arxiv.org/abs/2109.07958}, 
}

@inproceedings{min2023factscore,
    title = "{FA}ct{S}core: Fine-grained Atomic Evaluation of Factual Precision in Long Form Text Generation",
    author = "Min, Sewon  and
      Krishna, Kalpesh  and
      Lyu, Xinxi  and
      Lewis, Mike  and
      Yih, Wen-tau  and
      Koh, Pang  and
      Iyyer, Mohit  and
      Zettlemoyer, Luke  and
      Hajishirzi, Hannaneh",
    booktitle = "Proceedings of the 2023 Conference on Empirical Methods in Natural Language Processing",
    month = dec,
    year = "2023",
    publisher = "Association for Computational Linguistics",
    url = "https://aclanthology.org/2023.emnlp-main.741/",
    doi = "10.18653/v1/2023.emnlp-main.741",
    pages = "12076--12100",
}

@inproceedings{wang2024factuality,
    title = "Factuality of Large Language Models: A Survey",
    author = "Wang, Yuxia  and
      Wang, Minghan  and
      Manzoor, Muhammad Arslan  and
      Liu, Fei  and
      Georgiev, Georgi Nenkov  and
      Das, Rocktim Jyoti  and
      Nakov, Preslav",
    booktitle = "Proceedings of the 2024 Conference on Empirical Methods in Natural Language Processing",
    month = nov,
    year = "2024",
    publisher = "Association for Computational Linguistics",
    url = "https://aclanthology.org/2024.emnlp-main.1088/",
    doi = "10.18653/v1/2024.emnlp-main.1088",
    pages = "19519--19529",
}

@misc{wei2024simpleqa,
      title={Measuring short-form factuality in large language models}, 
      author={Jason Wei and Nguyen Karina and Hyung Won Chung and Yunxin Joy Jiao and Spencer Papay and Amelia Glaese and John Schulman and William Fedus},
      year={2024},
      eprint={2411.04368},
      archivePrefix={arXiv},
      primaryClass={cs.CL},
      url={https://arxiv.org/abs/2411.04368}, 
}

@inproceedings{yang2018hotpotqa,
    title = "{H}otpot{QA}: A Dataset for Diverse, Explainable Multi-hop Question Answering",
    author = "Yang, Zhilin  and
      Qi, Peng  and
      Zhang, Saizheng  and
      Bengio, Yoshua  and
      Cohen, William  and
      Salakhutdinov, Ruslan  and
      Manning, Christopher D.",
    booktitle = "Proceedings of the 2018 Conference on Empirical Methods in Natural Language Processing",
    month = oct # "-" # nov,
    year = "2018",
    address = "Brussels, Belgium",
    publisher = "Association for Computational Linguistics",
    url = "https://aclanthology.org/D18-1259/",
    doi = "10.18653/v1/D18-1259",
    pages = "2369--2380",
}

@misc{wang2025conflicting,
      title={Retrieval-Augmented Generation with Conflicting Evidence}, 
      author={Han Wang and Archiki Prasad and Elias Stengel-Eskin and Mohit Bansal},
      year={2025},
      eprint={2504.13079},
      archivePrefix={arXiv},
      primaryClass={cs.CL},
      url={https://arxiv.org/abs/2504.13079}, 
}

@misc{openai2025simpleevals,
  author = {{OpenAI}},
  title = {simple-evals},
  year = {2025},
  howpublished = {GitHub repository},
  note = {Archived benchmark results table and reference implementation; accessed March 12, 2026},
  url = {https://github.com/openai/simple-evals}
}

@misc{anthropic, 
    title={Introducing Claude Sonnet 4.6}, 
    url={https://www.anthropic.com/news/claude-sonnet-4-6}, 
    author={Anthropic}, 
    year={2026} 
}

@misc{openai,
    title={GPT-5 is here},
    url={https://openai.com/gpt-5/},
    author={OpenAI},
    year={2025} 
}

@inproceedings{xu1996localglobal_sigir,
    author = {Xu, Jinxi and Croft, W. Bruce},
    title = {Query expansion using local and global document analysis},
    year = {1996},
    isbn = {0897917928},
    publisher = {Association for Computing Machinery},
    url = {https://doi.org/10.1145/243199.243202},
    doi = {10.1145/243199.243202},
    booktitle = {Proceedings of the 19th Annual International ACM SIGIR Conference on Research and Development in Information Retrieval},
    pages = {4–11},
    series = {SIGIR '96}
}

@inproceedings{lavrenko2001rlm_sigir,
    author = {Lavrenko, Victor and Croft, W. Bruce},
    title = {Relevance based language models},
    year = {2001},
    isbn = {1581133316},
    publisher = {Association for Computing Machinery},
    url = {https://doi.org/10.1145/383952.383972},
    doi = {10.1145/383952.383972},
    booktitle = {Proceedings of the 24th Annual International ACM SIGIR Conference on Research and Development in Information Retrieval},
    pages = {120–127},
    series = {SIGIR '01}
}

@article{carpineto2012aqe_survey,
    author = {Carpineto, Claudio and Romano, Giovanni},
    title = {A Survey of Automatic Query Expansion in Information Retrieval},
    year = {2012},
    issue_date = {January 2012},
    publisher = {Association for Computing Machinery},
    volume = {44},
    number = {1},
    issn = {0360-0300},
    url = {https://doi.org/10.1145/2071389.2071390},
    doi = {10.1145/2071389.2071390},
    journal = {ACM Comput. Surv.},
    month = {jan},
    articleno = {1}
}

@misc{hyde2023acl_query,
      title={Precise Zero-Shot Dense Retrieval without Relevance Labels}, 
      author={Luyu Gao and Xueguang Ma and Jimmy Lin and Jamie Callan},
      year={2022},
      eprint={2212.10496},
      archivePrefix={arXiv},
      primaryClass={cs.IR},
      url={https://arxiv.org/abs/2212.10496}, 
}

@misc{query2doc2023emnlp_query,
      title={Query2doc: Query Expansion with Large Language Models}, 
      author={Liang Wang and Nan Yang and Furu Wei},
      year={2023},
      eprint={2303.07678},
      archivePrefix={arXiv},
      primaryClass={cs.IR},
      url={https://arxiv.org/abs/2303.07678}, 
}

\end{document}